# Implementation of AI Deep Learning Algorithm For Multi-Modal Sentiment Analysis

Jiazhen Wang

University of Wisconsin-Madison, Wisconsin, USA

jwang2477@wisc.edu

*Abstract*—A multi-modal emotion recognition method was established by combining two-channel convolutional neural network with ring network. This method can extract emotional information effectively and improve learning efficiency. The words were vectorized with GloVe, and the word vector was input into the convolutional neural network. Combining attention mechanism and maximum pool converter BiSRU channel, the local deep emotion and pre-post sequential emotion semantics are obtained. Finally, multiple features are fused and input as the polarity of emotion, so as to achieve the emotion analysis of the target. Experiments show that the emotion analysis method based on feature fusion can effectively improve the recognition accuracy of emotion data set and reduce the learning time. The model has a certain generalization.

## I. INTRODUCTION

Text emotion analysis refers to the analysis and processing of words such as "people's evaluation of goods, services, events and other entities" to obtain the subjective emotion information to be displayed. The research contents include: classification of emotion information, extraction of emotion information, emotion analysis and so on. At present, the commonly used emotion recognition techniques mainly include SVM, conditional random field, information entropy, etc., and they are all based on word bags. For example, some scholars [1] applied support vector machines to the emotion recognition and classification of sentences. However, this method tends to be sparse and high-dimensional for large-scale data. In recent years, domestic and foreign scholars [2] have successively launched new algorithms based on deep neural networks, opening up a new way for the research of the above problems. At present, many neural network methods are commonly used based on convolutional neural network (CNN), sequence basis (RNN) and tree structure (RAE). CNN, as used in reference [3], classifies the polarity of emotions. In literature [4], bidirectional sequence model (BLSTM) was used to study Chinese text classification. Because of its outstanding advantages in text feature extraction and sentiment analysis, it has attracted much attention from researchers in recent years. Literature [5] is a typical application example of recursive self-coding algorithm. This project intends to construct a multimodal emotion recognition method based on two pathways CNN and one bidirectional simple circuit cell (BiSRU). The method quantifies words using GloVe and captures keywords using CNN. Obtain the deep and meaningful emotional features contained in the text from a local perspective. Text context-dependent semantics are mined based on BiSRU's ability to analyze time series data. The overall temporal emotional features are extracted from the text to overcome the ability of CNN to process temporal data. This paper studies the automatic acquisition of word importance based on attention mechanism and combines it with the maximum pooling feature of BiSRU. Finally, the double-channel emotional features are integrated to fully mine the emotional information of the text and make the feature information more comprehensive.

## II. FEATURE FUSION EMOTION ANALYSIS MODEL

Vectorization of small samples is intended. Convolutional neural network is used for deep learning of small samples as the input layer of LSTM. The error back propagation algorithm is used to train the model [6]. Therefore, the feature fusion emotion analysis model can not only learn the local features of short microblog texts, but also learn the long-distance context history information.

### A. Text vectorization

The neural network takes the text vector as input and converts the text data into a one-dimensional real number vector. At present, there are two kinds of vectorized representation of text: primary representation and divergent representation. one-hot representation means that each word is represented by a large vector whose dimension is the same as the size of the vocabulary, which is usually extremely rare, and that no two words are associated with each other [7]. The distributed representation is represented in low-dimensional vectors, allowing related words to be semantically closer together, which is called "embedding". This article is subdivided into two representations: one hot and word embedding. 1 hot is to use a splitter to segment the word, and then perform one-hot encoding on the word to generate a "one hot" dictionary. Lexical embeddings use segmentation tools to segment words. word2vec is used to learn the vocabulary vector, and the dictionary of word embedding is eventually generated. Figure 1 shows the text Vectorization process algorithm (the picture is quoted in Vectorization Techniques in NLP [Guide]).

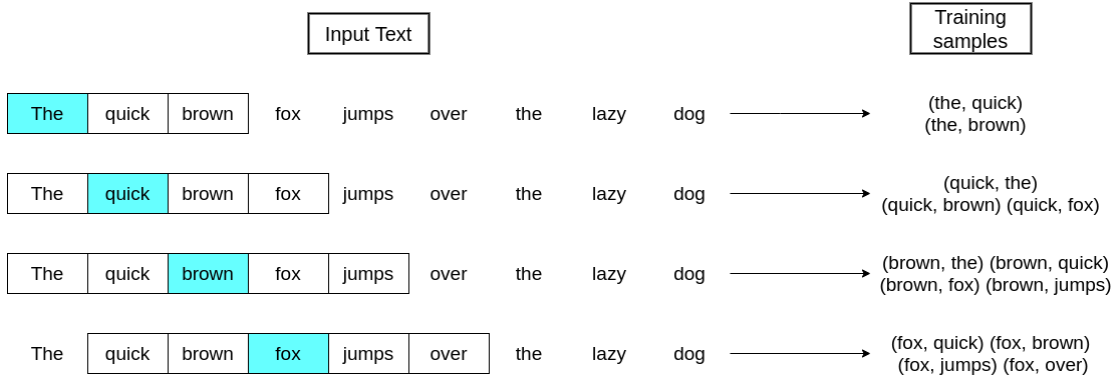

Fig. 1. Text vectorization process algorithm

### B. Emotion recognition method of convolutional neural network

Although CNN can extract some local features from text, it can't handle the long-term context correlation problem well. However, because short-term memory models can be learned over a long period of time, they can efficiently use a wide range of background knowledge [8]. In this paper, a novel method based on convolutional memory network is proposed for emotion recognition of text. As shown in Figure 2, the network layer of the feature fusion emotion analysis model includes convolution layer, pooling layer, timing layer and output layer (the picture is quoted in Accurate deep neural network inference using computational phase-change memory).

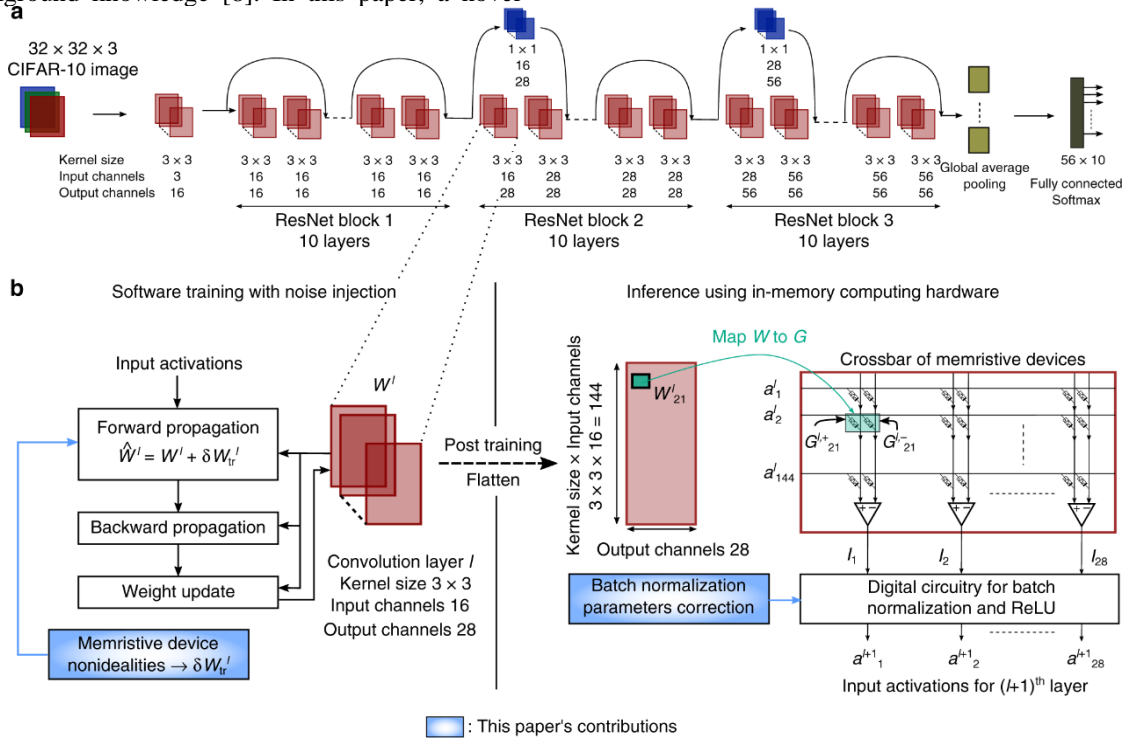

Fig. 2. Convolutional memory neural network model

The local character of text is extracted by convolutional network. In this algorithm, the original image is filtered first, and then the image is segmented by convolutional algorithm. In this paper, a convolution operation is proposed, which uses multiple convolution kernels of different sizes to construct new vectors. The row embedding algorithm is relearned to make it better use of the original data characteristics. Text is inserted before convolution, and then images are inserted [9]. Convolutional neural algorithm can extract small samples with specific meaning from small samples. Therefore, a multi-layer convolutional network is proposed to realize emotion extraction. After sampling the sample layer, the corresponding local features can be obtained.

### III. SEMI-SUPERVISED TEXT EMOTION ANALYSIS OF FAST LINK SYNTAX

#### A. Lexical vector represents vocabulary

Compared with existing vocabulary package theory, Semi-Supervised RAE uses word vector to represent vocabulary, for example, "college student" is represented by (0, 1, 0, 0). "Teacher" is represented by (1, 1, 0, 1). If A sentence $u$ contains $m$ words, then the $t$ word is $u_t$,. The word $u_t$ is mapped to the $n$ dimensional real vector space in the form of A standard normal distribution, which can be represented as $u_t \in R^n$. The word vectors for all words are stored in a word embedding matrix $S \in R^{n \times |Y|}$, where $Y$ is the size of the vocabulary. Then the vector for $u_t$ is

$$u_t = S\varepsilon_t \in R^n \quad (1)$$

$\varepsilon_t$ is a binary vector with a dimension of thesaurus size and a value of 0 or 1, all positions being 0 except the $t$ index.

### B. Guided loop automatic coding

Tree-based information is usually used to obtain the low-dimensional vector representation of the sentence, that is, supervised loop self-coding. Suppose $u = (u_1, u_2, \cdots, u_m)$ sentence is represented by a vector, and the input word vector of the child node $z_1, z_2$ is $u_1, u_2$, then the calculation method of the parent node $f_i$ is

$$f_i = g(w_{gi}Z) + \varepsilon_{gi} \quad (2)$$

$i = 1, 2, \cdots; Z = [z_1; z_2]$ represents the word vector concatenation matrix of $z_1, z_2$. $g(\cdot)$ is the activation function of the network. $w_{rj}$ and $\varepsilon_{rj}$ are the weight and bias parameters for calculating the parent node, respectively. A reconstruction layer of corresponding child nodes is added to each parent node to test the expressiveness of the parent node (Figure 3 is quoted in Deep Learning for High-Impedance Fault Detection: Convolutional Autoencoders). By calculating the difference between the initial node and the reconstructed node, the error of each node is calculated.

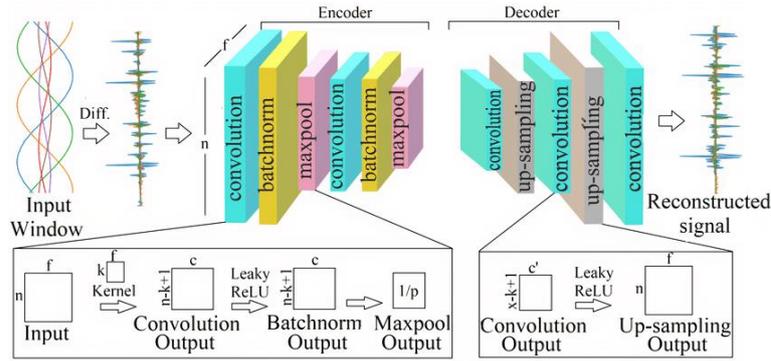

Fig. 3. Supervised recursive self-coding structure

The calculation method of reconstructed nodes is

$$Z' = w_{rj}f_i + \varepsilon_{rj} \quad (3)$$

Where: $j = 1, 2, \cdots; Z' = [z_1'; z_2']$ is the reconstruction node $z_1'$ of $z_1, z_2$. $z_2'$ word vector concatenation matrix. $\varepsilon_{rj}$ is the offset term. $w_{rj}$ is the weight parameter matrix. The word vector is reconstructed by using Euclidean distance calculation, namely

$$W_{rec}(Z) = \frac{1}{2} \| Z - Z' \|^2 \quad (4)$$

### C. Automatic coding of undirected loops

The tree of a general sentence is generally unknown. This paper presents a self-coding algorithm based on tree automatic learning. The optimization objective function of the tree structure prediction process is

$$R_\delta(u) = \arg\min_{v \in \lambda(u)} \sum_{d \in T(v)} W_{rec}(Z_d) \quad (5)$$

$R_\delta(u)$ is the optimal tree structure model of sentence $u$. θ is the parameter set. Set $\lambda(u)$ is the set of all possible tree structures of sentence $u$. $v$ is one of these structures. $d$ is a ternary structure with no terminal node in the calculation process. $Z_d = [z_{d1}, z_{d2}]; T(v)$ is the search function of this ternary structure. Since the degree of contribution of words to sentence meaning varies, each word should be weighted accordingly when calculating reconstruction errors.

$$W_{rec}(z_1, z_2, \delta) = \frac{n_1}{n_1 + n_2} \| z_1 - z_1' \|^2 + \frac{n_2}{n_1 + n_2} \| z_2 - z_2' \|^2 \quad (6)$$

$n_1, n_2$ is the number of words under the current child node $z_1, z_2$. In order to avoid obtaining too few parent nodes in the process of iterating repeatedly to reduce reconstruction errors, which brings inconvenience to the following operations, formula (2) is standardized here:

$$F = \frac{f_i}{\| f_i \|} \quad (7)$$

## D. Semi-supervised loop automatic coding

After obtaining the vector expression of the sentence, in order to estimate the emotional trend of the whole expression. Add a softmax (·) classifier to the network:

$$h = \text{soft}\max(\eta_l F) \quad (8)$$

$l$ is the current type of emotion. $\eta_l$ is the parameter matrix. If there is an $T$ emotion, then $h \in R^T$, and

$$W_{ce}(F, t, \delta) = -\sum_{t=1}^{T} t_t \log h_t(F, \delta) \quad (9)$$

$t$ indicates the label distribution. $t_t$ is the label distribution of the first $t$ emotion. $h_t$ is a conditional probability, and

$$h_t = f(t \mid Z) \quad (10)$$

The optimization objective function of semi-supervised recursion self-coding on the data set is

$$L = \frac{1}{N} \sum_{(u,t)} W(u, t, \delta) + \frac{\mu}{2} \|\delta\|^2 \quad (11)$$

$N$ is the training data set size. $\mu$ is the regular term coefficient of $S_2$.

$$W(u, t, \delta) = \sum_{d \in T(R_\delta(u))} W(Z_d, F_d, t, \delta) \quad (12)$$

$$W(Z_d, F_d, t, \delta) = \theta W_{rec}(Z_d, \delta) + (1-\theta) W_{ce}(F_d, t, \delta) \quad (13)$$

$\theta$ is the L-BFGS algorithm the optimal solution of the optimization objective function (11), where the gradient used is

$$\frac{\partial L}{\partial \delta} = \frac{1}{N} \sum_{(u,t)} \frac{\partial W(u, t; \delta)}{\partial \delta} + \mu \delta \quad (14)$$

In formula $\delta = \{\eta_{gi}, \varepsilon_{gi}, \eta_{ri}, \varepsilon_{rj}, \eta_l, S\}$.

## IV. EXPERIMENTAL RESULTS AND ANALYSIS

### A. Experimental data set

The paper used four groups of English public emotion samples. 1) MR Is a double-category table of emotions in English film reviews, including two dimensions of positive and negative emotions. 2) CR refers to a user's evaluation of different products, which is a data set of two categories of emotion, respectively negative and positive. 3) SST-2 divides the film into two categories, namely: training, confirmation and test, among which the emotional category has negative category and positive category. 4) Subj is a set of subjective evaluations with subjective evaluations and objective markers. Unsegmented training samples, test samples, magnetic resonance and CR samples of test samples are given in this paper. The test was carried out by cross-validation method. Table 1 shows information about the size of the data set.

TABLE I. DATA SET SIZE INFORMATION

| Data set | MR | CR | SST-2 | Subj |
|---|---|---|---|---|
| Sample size/piece | 11106 | 3932 | 10014 | 10417 |
| Average sentence length | 21 | 20 | 20 | 24 |
| Test set | CV | CV | 1821 | CV |

### B. Comparison experiment Settings

The different dimension of word vector has certain influence on the recognition result. This thesis firstly makes a comparative study on the vector scale of vocabulary. Word vectorization was performed using a 50-dimensional glove.6B.50D, a 100-dimensional glove.6B.100d, a 200-dimensional glove.6B.200D and a 300-dimensional glove.6B.300d, respectively. The convolutional neural network model is tested using MR Data. In order to study the classification, the most suitable word vector dimension is obtained. The resulting effect is shown in Figure 4. The results show that the prediction accuracy of the convolutional neural network algorithm is the highest when the word vector dimension is 300.

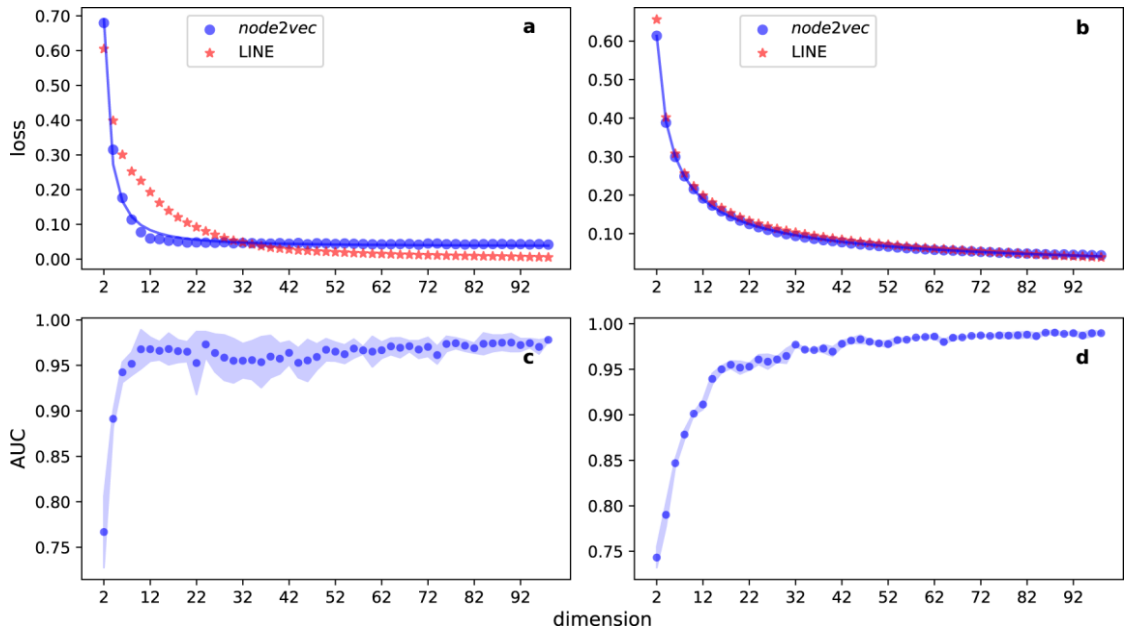

Fig. 4. Comparison of word vector dimensions

The experiment was conducted on four different sets of emotion words. In order to verify the validity of the convolutional neural network model proposed in this paper, the proposed model is compared with some traditional neural network models.

*C. Experimental study and conclusion*

Convolutional neural networks were compared with 5 different training modes. Experimental comparison results are shown in Table 2. Experiments show that the convolutional neural network model proposed in this paper has better classification accuracy than the other five models on the four data sets, and the effectiveness of this method in text emotion recognition is verified by experiments.

TABLE II. MODEL COMPARISON RESULTS

| Model | MR | CR | SST-2 | Subj |
|---|---|---|---|---|
| Kim CNN | 82.00 | 84.94 | 86.77 | 96.78 |
| BiLSTM | 81.66 | 84.44 | 86.09 | 97.25 |
| CNN-BiLSTM | 82.27 | 85.18 | 87.06 | 97.78 |
| CNN-BiSRU | 83.22 | 85.59 | 88.03 | 97.55 |
| CNN-BiLSTM-MA | 82.85 | 86.43 | 86.21 | 97.47 |
| Convolutional neural network | 85.08 | 88.57 | 89.81 | 98.09 |

By comparing the training time consumed by five different types of neural networks on different samples, and comparing their performance on SST-2. It can be seen from Table 3 that the computational speed of the qubit realized by the proposed method is only 340 milliseconds, which is much lower than that of the bilinear short-term memory algorithm. The experimental results show that this method can realize parallel processing of text and reduce the time required for learning.

TABLE III. TRAINING TIME

| Model | Training time/ms |
|---|---|
| Kim CNN | 323 |
| BiLSTM | 917 |
| CNN-BiLSTM | 1000 |
| CNN-BiSRU | 302 |
| CNN-BiLSTM-MA | 1042 |
| Convolutional neural network | 354 |

## V. CONCLUSION

A multimodal emotion detection method based on convolutional network and bidirectional single loop unit is proposed. Convolutional neural network is used to extract context-dependent semantics, and the attention is fused with the maximum pooled bidirectional simple loop to achieve the effective fusion of context-dependent semantic information. In order to obtain more rich emotional characteristics. This improves the performance of emotion recognition and speeds up the learning process. The effectiveness of the proposed method is proved by comparison with several classical neural networks.

REFERENCES

[1] Li Xiaoyan, FU Huitong, NIU Wentao, WANG Peng, LV Zhigang, WANG Weiming. Multi-modal pedestrian detection algorithm based on deep learning. Journal of Xi 'an Jiaotong University, vol. 56, pp. 61-70,October 2021.

[2] Zhou Xiangzhen, LI Shuai, SUI Dong. Weibo emotion analysis based on deep learning and attention mechanism. Journal of Nanjing Normal University: Natural Science Edition, vol.46, pp. 115-121,February 2022.

[3] Li Yang, GUO Yuzhe, PANG Le. Research on intelligent analysis and verification algorithm of medical data based on deep learning. Electronic Design Engineering, vol. 31, pp. 35-39,July 2022.

[4] Yang Lingling. Speech enhancement deep learning algorithm based on joint loss function. Electronic Products World, vol. 30(6) : 75-77,June 2022.

[5] Liu Fangtao, Chang Rui, Liu Cui Shi, et al. Research on image quality volume model of deep learning reconstruction algorithm. Theoretical and Applied Research of CT, vol. 31, pp.62-66,March 2022.

[6] Guan Shaoya, Zhang Cheng, Meng CAI, et al. Vascular ultrasound image segmentation algorithm based on phase symmetry. Journal of

Beijing University of Aeronautics and Astronautics, vol. 49, pp. 2645-2650,October 2022.

[7] Real-time deep learning tracking algorithm based on NPU. Journal of Applied Optics, vol.43, pp.110-114,April 2022.

[8] Wang Jinzhu. Identity recognition algorithm based on deep learning and gait analysis. Electronic Design Engineering, vol.30, pp. 15-22,July 2021.

[9] Wang Chuan-Yu, LI Wei-xiang, Chen Zhen-huan. Multimodal emotion recognition based on speech and video images. Computer Engineering and Applications, vol. 57, pp.86-92,March 2021.